\documentclass[10pt,twocolumn,letterpaper]{article}
\usepackage[pagenumbers]{cvpr}
\usepackage[normalem]{ulem}
\usepackage{algorithm}
\usepackage{algpseudocode}

\algrenewcommand\algorithmiccomment[1]{\hfill{\footnotesize$\triangleright$ #1}}
\usepackage{multirow}
\newcommand{\Top}[1]{\operatorname{Top}_{#1}}   
\DeclareMathOperator{\softmax}{softmax}

\usepackage{colortbl}
\usepackage{tabularx}
\definecolor{paperTableHighlight}{HTML}{EDF3F8}
\newcommand{\resulttablefont}{\fontencoding{T1}\fontfamily{ptm}\fontsize{9}{11}\selectfont}

\newcommand{\Dense}{Dense}
\newcommand{\ExactTopK}{Exact Top-\ensuremath{k}}

\makeatletter
\def\supp@floatcommands{@dbflt,@dblfloatplacement,@addtodblcol,@addtocurcol,@addtotoporbot,@addtobot,@doclearpage,@combinedblfloats}
\@for\supp@cmd:=\supp@floatcommands\do{%
  \expandafter\let\csname supp@original@\supp@cmd\expandafter\endcsname\csname\supp@cmd\endcsname}
\makeatother
\usepackage{stfloats}
\makeatletter
\@for\supp@cmd:=\supp@floatcommands\do{%
  \expandafter\let\csname supp@enabled@\supp@cmd\expandafter\endcsname\csname\supp@cmd\endcsname
  \expandafter\let\csname\supp@cmd\expandafter\endcsname\csname supp@original@\supp@cmd\endcsname}
\newcommand{\EnableSupplementaryFloats}{%
  \@for\supp@cmd:=\supp@floatcommands\do{%
    \expandafter\let\csname\supp@cmd\expandafter\endcsname\csname supp@enabled@\supp@cmd\endcsname}}
\makeatother

\makeatletter
\renewcommand\paragraph{\@startsection{paragraph}{4}{\z@}%
  {8pt plus 1pt minus 1pt}{-1em}{\normalfont\normalsize\bfseries}}
\makeatother

\definecolor{cvprblue}{rgb}{0.21,0.49,0.74}
\usepackage[breaklinks,colorlinks,allcolors=cvprblue]{hyperref}
\newcommand{\name}{TaRo}
\title{{\name}: Temporal Reuse of Top-Ranked KV Cache Entries for Efficient Streaming Video-Language Models}
\author{Soomin Lee \quad Omkar Prabhune \quad Arghadip Das \quad Vijay Raghunathan \quad Younghyun Kim\\
Purdue University\\
{\tt\small \{lee5782,oprabhun,das169,vr,younghyun\}@purdue.edu}}
\hypersetup{pdfauthor={Soomin Lee, Omkar Prabhune, Arghadip Das, Vijay Raghunathan, Younghyun Kim}}

\hypersetup{pdftitle={TaRo: Temporal Reuse of Top-Ranked KV Cache Entries for Efficient Streaming Video-Language Models}}
\begin{document}
\maketitle
\begin{abstract}
Selective attention reduces the number of key--value (KV) entries used by streaming video-language models, but finding those entries can still require repeated search over a growing history.
As frames accumulate, both absolute latency and its increase with history length matter for streaming efficiency.
We propose {\name}, a training-free method that reuses KV candidate banks at corresponding visual-token positions across frames while retaining the complete KV history.
The current query rescores bank candidates together with the current frame's causal keys to select the top-$k$ entries for attention.
Full-history search refreshes a bank when query similarity falls below a threshold or the bank reaches an age limit.
Evaluated on two models using subsets of OVO-Bench and StreamingBench, {\name} achieves 1.37$\times$ to 3.58$\times$ LLM-prefill speedups over ReKV, with accuracy changes ranging from $-$4.60 to +6.59 percentage points.
Compared with {\ExactTopK}, {\name} also exhibits smaller latency growth as KV history increases.
\end{abstract}

\begin{figure}[!t]
  \centering
  \includegraphics[width=\columnwidth]{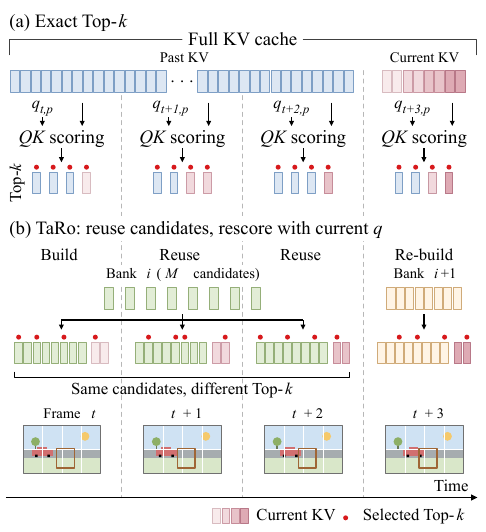}
  \caption{Temporal candidate reuse at a fixed visual-token position $p$ ($M=8$, $k=4$). (a) {\ExactTopK} searches the full causal history for each query. (b) {\name} rescores a reused bank and current causal KV, allowing the selected Top-$k$ to change between refreshes.}
  \label{fig:concept}
\end{figure}

\section{Introduction}
\label{sec:intro}

Streaming video-language models (VideoLLMs) must process video while new frames continue to arrive, enabling timely responses for live assistants, wearable cameras, and robots~\cite{chen2024videollmonline,qian2025dispider,zhang2025flashvstream}.
As frames accumulate in the language model's key--value (KV) cache, each new frame attends to a longer history, increasing the cost of keeping up with the stream~\cite{di2025rekv,xu2026streamingvlm,kim2026vrex}.

\begin{figure*}[!t]
  \centering
  \includegraphics[width=\textwidth]{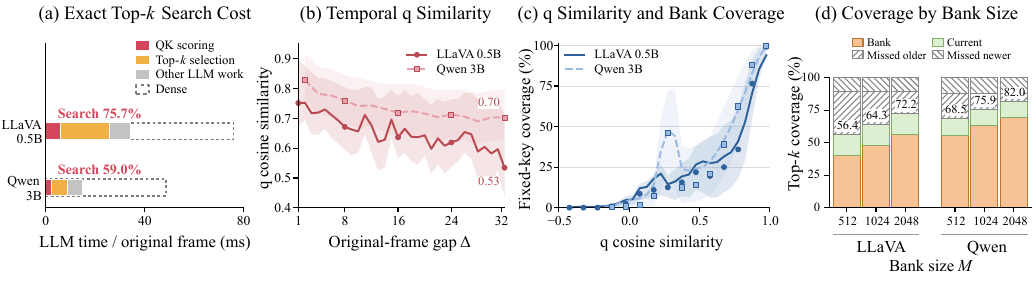}
  \caption{(a) {\ExactTopK} search cost ($k=128$). (b) Temporal Q similarity. (c) Prior-bank coverage of fixed-key top-128; lines: means, markers: medians. (d) Full-history top-128 coverage by bank+current KV. Shading in (b,c): IQR.
 }
  \label{fig:motivation}
\end{figure*}

Standard attention scores every causally accessible KV entry; we refer to it as {\Dense}.
Selective attention restricts each query to a subset of these entries~\cite{tang2024quest,xiao2024infllm,wu2025tokenselect}.
We denote by {\ExactTopK} the variant that scores the full causal key history and attends only to its $k$ highest-scoring entries.
Although {\ExactTopK} uses only $k$ entries for the attention output, finding them still requires a full-history search at every processing step.
Our profiling shows that this search consumes a substantial share of LLM time.

Streaming video offers an opportunity to avoid repeating this search.
Queries at corresponding visual-token positions remain similar across nearby frames, and on a fixed key set, higher query similarity is generally associated with better coverage of the current top-$k$ selection by earlier candidates.
These observations motivate reusing candidates across nearby frames while checking when they need to be refreshed.

We propose {\name}, a training-free method that maintains candidate KV indices for each layer, query head, and visual-token position as shown in~\cref{fig:concept}.
The current query rescores these candidates together with the current frame's causal keys, so the final attention set can change without a full-history search.
Between refreshes, each step scores a bounded candidate set whose size does not grow with the history.
A full-history search is triggered when the current query differs sufficiently from the reference query saved at the last refresh, or when the bank reaches its age limit. The complete KV history is retained throughout.

Our key contributions are as follows.
\begin{itemize}
    \item \textbf{An empirical basis for temporal candidate reuse.}
    We show that full-history key scoring and top-$k$ selection remain a substantial bottleneck in streaming video prefill.
    We further find that queries at corresponding visual-token positions exhibit temporal similarity, and that higher query similarity is associated with better coverage of the current top-$k$ keys by previously selected candidates.
    
  \item \textbf{Position-wise KV candidate reuse.}
  We introduce a candidate-bank design that exploits the correspondence between visual-token positions across successive frames.
  Each bank stores the indices of past KV entries selected for the corresponding spatial location.
  For an incoming frame, {\name} combines these candidates with the current frame's causal keys and rescores them with the current query to select the final attention set.
  This separates candidate discovery from per-frame selection, so the full-history search no longer runs at every frame.

  \item \textbf{Query-guided refresh of position-wise banks.}
  We develop a refresh policy that compares the current query with the reference query recorded when the bank was last rebuilt.
  A sufficiently large query change or an age limit triggers a full-history search that rebuilds the bank. Otherwise, {\name} continues to select from the existing candidates and current-frame keys.
  Our fixed-key analysis relates query similarity to candidate coverage, supporting the use of query similarity as a refresh signal, though not as a guarantee of coverage. 
  
\end{itemize}

\section{Background and Motivation}
\label{sec:motivation}

\subsection{Background}
\label{sec:background}

A streaming VideoLLM encodes incoming frames into visual tokens and prefills them into its language model, appending their keys and values to the KV cache.
Frames are prefilled in processing units of one or more original frames, and each visual-token query can attend to earlier tokens and to tokens up to its own position in the current processing unit under the causal mask.
As the stream continues, the causal history available to each new query grows.

{\Dense} scores every accessible key and combines the corresponding values using softmax weights.
{\ExactTopK} also scores every accessible key, but selects only the $k$ highest-scoring entries before applying softmax and value aggregation~\cite{gupta2021topk}.
For fixed $k$, attention over the selected entries uses a fixed number of values, whereas finding those entries still searches the growing history.
Reducing the attended set therefore does not eliminate the repeated search.We next measure the cost of this search and examine whether candidates can be reused across frames.

\subsection{Motivation}
\label{sec:motivation_study}

We organize our study around three observations: (\textbf{O1}) search remains costly, (\textbf{O2}) queries exhibit temporal similarity, and (\textbf{O3}) query similarity is associated with candidate validity.
We study LLaVA-OneVision-0.5B~\cite{li2024llavaonevision} and Qwen2.5-VL-3B~\cite{bai2025qwen25vl} on videos from OVO-Bench~\cite{niu2025ovobench} and StreamingBench~\cite{lin2026streamingbench}.

\paragraph{O1. Full-history search remains costly.}
\label{sec:study}
Even after restricting value aggregation to 128 entries, searching for those entries accounts for a substantial share of LLM time.
We profile LLM prefill over original frames 119--126, corresponding to approximately 24K KV entries for LLaVA and 12K for Qwen.
Fig.~\ref{fig:motivation}(a) shows the time breakdown for optimized {\ExactTopK} with $k=128$.
QK scoring and top-$k$ selection take 6.2 and 19.7\,ms per original frame for LLaVA, together accounting for 75.7\% of its 34.2\,ms profiled LLM time.
For Qwen, the corresponding costs are 2.4 and 6.5\,ms, or 59.0\% of 15.1\,ms.
Thus, shrinking the attended set leaves search as a major cost, motivating methods that avoid repeating it at every step.

\paragraph{O2. Queries exhibit temporal similarity.}
\label{sec:q_temporal}
Queries are collected from {\Dense} runs. Post-RoPE queries at corresponding visual-token positions within each layer and query head are compared in Fig.~\ref{fig:motivation}(b).
For every eligible reference unit, we compare its query with later queries up to a gap of 32 original frames.
We average cosines within each video and then weight videos equally.
Shading shows the video-balanced query interquartile range.

 Both models show high mean cosine similarity between adjacent units, 0.752 for LLaVA and 0.829 for Qwen.
 Even at the largest evaluated gap ($\Delta=32$), it remains 0.534 and 0.703, respectively.
This similarity strongly motivates reuse across nearby frames, while its decline with distance motivates checking whether earlier candidates remain useful.
Post-RoPE cosine includes positional rotation, so these changes cannot be attributed solely to scene content.
Query similarity alone also does not establish candidate validity, and O3 examines that connection directly.

\paragraph{O3. Query similarity is associated with candidate validity.}
\label{sec:q_motivation}
We test whether similar queries select keys that an earlier candidate bank can cover in Fig.~\ref{fig:motivation}(c).
At each layer, head, and visual-token position, the reference query selects a top-512 bank from its causally accessible keys.
The query from the next processing unit then rescores \emph{the same fixed key set} to obtain an exact top-128.
Coverage is the fraction of those 128 indices present in the saved bank.
Holding the keys fixed excludes misses due to newly available keys, isolating ranking changes associated with the changed query.
Coverage is generally higher at high Q cosine similarity, reaching mean values of 94.0\% for LLaVA and 99.1\% for Qwen at similarity $\geq 0.95$.
This association supports using Q cosine similarity as a refresh signal.

Coverage of the current full-history exact top-128 by a prior bank plus current causal KV is shown in Fig.~\ref{fig:motivation}(d).
\emph{Missed older} and \emph{Missed newer} denote uncovered keys that were causally available and unavailable to the reference query, respectively.
Increasing the bank size from 512 to 2048 raises coverage from 56.4\% to 72.2\% for LLaVA and from 68.5\% to 82.0\% for Qwen.
The remaining misses motivate bank refresh.

Together, these observations motivate candidate reuse with refresh: nearby queries offer reuse opportunities, Q similarity provides a signal of candidate validity, and missing candidates show why the bank needs to be refreshed.

\begin{figure*}[t]
  \centering
  \includegraphics[width=\textwidth]{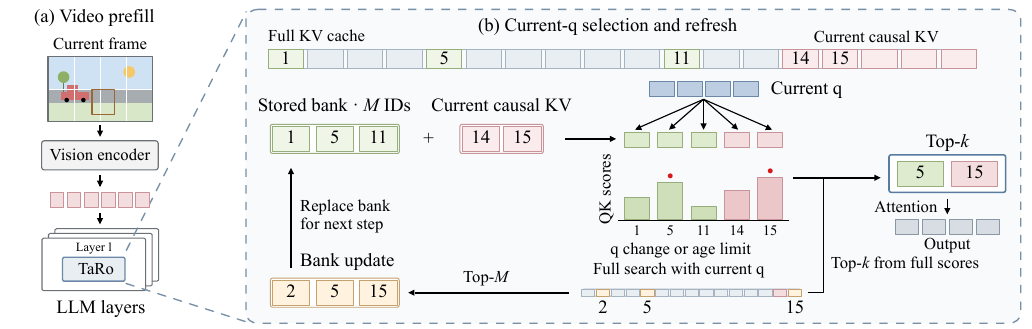}
  \caption{{\name} in streaming prefill. (a) Placement within the model. (b) Position-specific banks supply candidates for current-query rescoring and top-$k$ attention; Q change or bank age triggers full-history refresh. The example uses $M=3$ and $k=2$; red dots mark selected top-$k$ entries. Tokens, key IDs and scores are schematic.}
  \label{fig:overview}
\end{figure*}

\section{Related Work}
\label{sec:related}

\paragraph{Streaming VideoLLMs.}
Long-video and streaming VideoLLMs condense past frames into a compact memory~\cite{song2024moviechat,he2024malmm,qian2024videostreaming,zhang2025flashvstream} or learn when to respond within a stream~\cite{chen2024videollmonline,qian2025dispider}.
ReKV~\cite{di2025rekv} encodes frames with sliding-window attention, offloads the full video KV cache, and retrieves entries relevant to each question.
StreamingVLM~\cite{xu2026streamingvlm} keeps attention sinks, a short window of recent vision tokens, and a longer window of recent text tokens.
OVO-Bench~\cite{niu2025ovobench} and StreamingBench~\cite{lin2026streamingbench} evaluate online video understanding.
These systems address streaming memory and interaction; {\name} focuses on reducing repeated KV search during frame prefill.

\paragraph{KV Cache Pruning and Compression.}
StreamingLLM~\cite{xiao2024streamingllm} retains attention sinks and a recent window, H$_2$O~\cite{zhang2023h2o} keeps recent entries and those with high cumulative attention, and SnapKV~\cite{li2024snapkv} selects important KV positions for each attention head.
For streaming video, InfiniPot-V~\cite{kim2025infinipotv} prunes tokens using temporal redundancy and value-based importance, while StreamMem~\cite{yang2025streammem} combines query-agnostic selection with weighted KV merging.
These methods reduce the cache by removing or combining entries; {\name} preserves the complete KV history and reduces repeated search through candidate reuse.

\paragraph{Query-aware sparse attention.}
Top-$k$ attention~\cite{gupta2021topk} restricts each query to its highest-scoring keys; our full-history {\ExactTopK} baseline finds this subset by scoring every causally accessible key.
Dynamic sparse methods use page/block selection or structured sparsity~\cite{tang2024quest,xiao2024infllm,jiang2024minference,xu2025xattention}, low-cost score approximations~\cite{ribar2024sparq,singhania2024loki}, indexed retrieval for cross-attention or decoding~\cite{bertsch2023unlimiformer,liu2025retrievalattention}, or LSH-based attention sampling~\cite{chen2025magicpig}.
TokenSelect uses token-level selection with per-head soft voting~\cite{wu2025tokenselect}, NSA trains sparse attention natively~\cite{yuan2025nsa}, and LServe combines static and dynamic sparsity~\cite{yang2025lserve}.
{\name} targets the frequency of candidate discovery across video frames.

\paragraph{Reusing selections.}
Closest to {\name} are methods that reuse a selection instead of recomputing it.
During decoding, TokenSelect reuses cached selections for similar queries~\cite{wu2025tokenselect}, and RefreshKV preserves the full KV cache while using query similarity to decide when full attention should rebuild a smaller cache~\cite{xu2025refreshkv}.
HShare shares critical indices across queries, layers and heads~\cite{wu2025hshare}, while TidalDecode and OmniKV reuse selections across layers~\cite{yang2025tidaldecode,hao2025omnikv}.
For streaming video prefill, V-Rex combines spatial--temporal key clustering with dynamic retrieval and a hardware accelerator~\cite{kim2026vrex}.
{\name} reuses position-specific top-$M$ candidate banks across frames; each current query rescores its bank together with current causal keys to choose the final top-$k$ subset.

\paragraph{Temporal redundancy in video models.}
DeltaCNN~\cite{parger2022deltacnn} and Eventful Transformers~\cite{dutson2023eventful} reuse earlier computations to avoid recomputing unchanged regions or tokens across frames.
Token merging and pruning methods reduce visual sequence lengths in vision and vision-language models~\cite{bolya2023tome,chen2024fastv,yang2025visionzip,tao2025dycoke,yao2025timechatonline}.
{\name} targets repeated search while retaining the complete KV history.
These approaches are potentially complementary when a consistent correspondence between visual-token positions across frames is maintained.


\section{{\name}: Temporal Candidate Reuse}
\label{sec:method}

{\name} is a training-free method that reduces the repeated full-history search of {\ExactTopK} during streaming video prefill (\cref{fig:overview}).
It maintains a small set of candidate key indices at corresponding visual-token positions to enable temporal reuse across frames.
The current query rescores these candidates together with the causally accessible keys of the current frame, then selects the keys used for attention.
When the query changes sufficiently or the bank reaches an age bound, a full-history search rebuilds the candidates.
{\name} leaves the model weights unchanged and preserves the complete KV cache; it changes how keys are searched and selected, while retaining the usual softmax and weighted value sum over the selected keys.

\subsection{Candidate Bank Construction}
\label{sec:bank}
\label{sec:refresh_step}

For every layer $l$, query head $h$ and visual-token position $p$, {\name} maintains a candidate bank $\mathcal{B}_{l,h,p}$ of up to $M \ge k$ key indices.
To exploit temporal similarity at the same spatial location, positions correspond across frames through their index $p$ on a fixed visual-token grid.
{\name} operates within each LLM layer during video-frame prefill, as illustrated in \cref{fig:overview}(a); question and answer tokens use {\Dense}.

We index frames by $t$.
Each frame contains $P$ visual tokens; $T_t$ denotes the KV-cache length after appending frame $t$.
The causally accessible key indices for visual-token position $p$ are
\[
\mathcal A_{t,p}=\{0,\ldots,T_t-P+p\},
\qquad 0\le p<P.
\]
{\name} builds and refreshes banks through full-history search.
Let $t_r$ denote the frame of the most recent such search.
Using the query $q$ for $(l,h,p)$ at frame $t_r$, {\name} selects up to $M$ keys from $\mathcal A_{t_r,p}$ by their scaled dot-product scores and stores their indices:
\begin{equation}
s_j=\frac{q^\top k_j}{\sqrt d},
\qquad
\mathcal B_{l,h,p}
\leftarrow
\Top{M}\{s_j:j\in\mathcal A_{t_r,p}\},
\label{eq:refresh}
\end{equation}
where $k_j$ is the key at cache index $j$ and $d$ is the attention-head dimension.

A bank stores indices only, not attention scores or weights.
The query that constructed it is stored separately as the reference query, together with the last refresh time.
The first frame initializes every bank; when fewer than $M$ keys are accessible, all are stored.
For $n$ input scores and a requested selection size $b$, $\Top{b}$ returns the indices of the $\min(b,n)$ highest scores.
The example in \cref{fig:overview}(b) uses $M=3$ and $k=2$: the stored bank contains historical key indices $\{1,5,11\}$, whose corresponding KV entries remain in the complete cache.
The complete cache remains available for subsequent full-history searches.

\subsection{Candidate Reuse and Rescoring}
\label{sec:reuse}

When no refresh is required at frame $t$, {\name} searches the saved bank together with the current frame's causal key indices $\mathcal{F}_{t,p}$.
Their union defines the candidate index set $\mathcal{C}_{t,p}$:
\begin{equation}
  \begin{aligned}
    \mathcal{F}_{t,p}&=\{T_t-P,\ldots,T_t-P+p\},\\
    \mathcal{C}_{t,p}&=\mathcal{B}_{l,h,p}\cup\mathcal{F}_{t,p}.
  \end{aligned}
  \label{eq:candidates}
\end{equation}
Layer and head indices are implicit in $\mathcal{C}_{t,p}$.

The current query $q$ rescores the keys in $\mathcal{C}_{t,p}$ using the scaled dot product in \cref{eq:refresh} and selects the attention indices $\mathcal{S}=\Top{k}\{s_j:j\in\mathcal{C}_{t,p}\}$.
Let $s_{\mathcal{S}}$ denote the vector of scores indexed by $\mathcal{S}$, and $v_j$ the value at cache index $j$.
The attention output $o$ is
\begin{equation}
  o=\sum_{j\in\mathcal{S}}\softmax(s_{\mathcal{S}})_j\,v_j.
  \label{eq:attention}
\end{equation}
In \cref{fig:overview}(b), the stored indices $\{1,5,11\}$ are combined with the current frame's causal indices $\{14,15\}$.
Rescoring these five candidates with the current query selects $\{5,15\}$ for attention.
Thus, reusing a bank does not fix the final attention set: selection depends on the current query and can include newly available keys.
Keys appended between $t_r$ and $t$ remain in the full KV cache but are excluded from reuse candidates; they become eligible for selection at the next full-history search.

\subsection{Query-Similarity-Guided Refresh}
\label{sec:refresh}

The association in \cref{sec:q_motivation} motivates a query-similarity refresh policy for position-specific banks.
Related selection-reuse and cache-refresh mechanisms appear in TokenSelect and RefreshKV~\cite{wu2025tokenselect,xu2025refreshkv}.

Under grouped-query attention (GQA)~\cite{ainslie2023gqa}, let $\mathcal{G}$ be the query heads sharing one KV head at a fixed layer $l$ and position $p$.
Each head $h\in\mathcal{G}$ maintains its own bank $\mathcal{B}_h=\mathcal{B}_{l,h,p}$ and reference query $q_h^{\mathrm{ref}}$ saved at $t_r$.
The group shares a refresh decision and the timestamp $t_r$.
We write $q_h$ for the current query of head $h$ and use post-RoPE queries in the cosine-similarity test.

Let $\tau$ be the similarity threshold and $N_{\max}$ the maximum bank age in frames.
For existing banks, the group refreshes when
\begin{equation}
  \min_{h\in\mathcal{G}}\cos(q_h,q_h^{\mathrm{ref}})<\tau
  \quad\text{or}\quad t-t_r\ge N_{\max}.
  \label{eq:policy}
\end{equation}
A similarity below $\tau$ in any head triggers a refresh for the group.
The age bound ensures that keys from intervening frames are reconsidered even when queries remain similar.

On refresh, each head scores the full causal history with the current query, using these scores both to rebuild its top-$M$ bank and to select its top-$k$ attention set.
The lower path in \cref{fig:overview}(b) illustrates this update: the bank is replaced with $\{2,5,15\}$ for subsequent frames, while the current frame's attention selection also uses the full-history scores rather than being restricted to the previous bank.
Each reference query is updated to $q_h$, and the group records $t_r\leftarrow t$.
Between refreshes, the banks and reference queries are retained.
Decisions are independent across layers, KV-head groups and visual-token positions.

\subsection{Search Cost and Memory Overhead}
\label{sec:cost}

At a fixed cache length $T_t$ and position $p$, a full-history search scores $T_t-P+p+1$ keys per query head, whereas reuse scores at most $M+p+1$ keys.
For a refresh fraction $\phi$ among noninitial decisions, the mean number of scored keys per head is at most
\begin{equation}
  \phi\,(T_t-P+p+1)+(1-\phi)(M+p+1).
  \label{eq:search_cost}
\end{equation}
For a model with $L$ layers and $H$ query heads per layer, banks and reference queries store up to $LHPM$ indices and $LHPd$ values, respectively.
One refresh timestamp is stored per layer, KV-head group and position.
This additional state is independent of video length for fixed model dimensions, $P$ and $M$, while the full KV cache grows with $T_t$.

\section{Experimental Setup}
\label{sec:setup}

\paragraph{Models.}
We evaluate {\name} on LLaVA-OneVision-0.5B~\cite{li2024llavaonevision} and Qwen2.5-VL-3B~\cite{bai2025qwen25vl}.
In our experimental configuration, LLaVA converts each original frame into 196 visual tokens, whereas Qwen processes two original frames together as 196 visual tokens.
Each indexed frame $t$ in the method therefore corresponds to one original frame for LLaVA and a pair of original frames for Qwen.

\paragraph{Benchmarks.}
We use fixed evaluation subsets constructed from OVO-Bench~\cite{niu2025ovobench} and StreamingBench~\cite{lin2026streamingbench}.
The final subsets contain 182 questions from 126 videos for OVO-Bench and 87 questions from 74 videos for StreamingBench.

To evaluate repeated search costs and the effect of candidate reuse with large accumulated KV histories, we consider questions whose video-prefix KV length is at least 12K in LLaVA.
To process each video prefix and question in full, we apply a common input limit matching LLaVA's maximum context length of 32,768 tokens.
We exclude questions for which the combined length of the video prefix, question, answer options, and reserved space for answer generation exceeds this limit.
We also exclude questions whose timestamps or four-choice format do not meet the requirements of the evaluation protocol, videos with a source frame rate below 2 FPS, and questions whose required prefix exceeds the actual video duration.

We sample videos at 2 FPS and feed each prefix in chronological order up to its question timestamp.
All compared methods and both models use the same questions and original frames.
The 12K condition is defined using LLaVA's actual KV length; the same questions are retained for Qwen even when its actual KV length is below 12K.

\paragraph{Motivational analyses.}
Search-cost profiling uses twelve videos, six from OVO-Bench and six from StreamingBench.
Query-similarity and candidate-coverage analyses share twelve videos, six from each benchmark, and use sequential {\Dense} reference trajectories with synthetic terminal padding excluded.
For the query-similarity lag study, original-frame gaps are $\Delta=1,\ldots,32$ for LLaVA and $\Delta=2,4,\ldots,32$ for Qwen.

\paragraph{Baselines.}
We use {\ExactTopK} and ReKV~\cite{di2025rekv} as efficiency baselines and include {\Dense} as a reference.
{\Dense} attends to all causal KV entries accessible to each query.
{\ExactTopK} scores the keys in the entire causal history using the current query, then attends to the $k$ highest-scoring KV entries.
ReKV uses its official implementation and model-specific adapters, including its KV management and offload paths.
{\ExactTopK} and {\name} use the same final attention budget of $k=128$ in all reported comparisons.

\paragraph{{\name} configuration.}
We set the candidate-bank size to $M=512$ and the final attention budget to $k=128$.
Refresh decisions use the cosine similarity of post-RoPE queries, with thresholds of $\tau=0$ for LLaVA and $\tau=0.5$ for Qwen.
The maximum bank age is 32 original frames, corresponding to 32 steps for LLaVA and 16 steps for Qwen.
For each model, we use the same configuration for the main accuracy and latency evaluations.
Threshold sweeps and comparisons with Fixed-Interval Refresh are conducted separately.

\paragraph{Metrics.}
Multiple-choice accuracy is the fraction of evaluation questions for which the option letter (A--D) extracted from the model's generated response matches the correct option.
Responses from which no valid option letter can be extracted are counted as incorrect.

We report latency in milliseconds per original frame and separately measure LLM latency and Vision+LLM latency, which also includes the vision encoder.
The main table reports LLM latency.
We group LLM latency measurements by actual KV length: 0--8K, 8--16K, 16--24K, 24--30K, and 30K--end for LLaVA; 0--4K, 4--8K, 8--12K, 12--15K, and 15K--end for Qwen.
Here K denotes 1,000 entries; intervals include their lower bound and exclude their upper bound, with the last interval extending to the end of the evaluated prefix.
Accuracy uses all evaluation questions. For latency, we retain the longest evaluated prefix per source video, breaking ties by question ID, and use the same selected prefixes for all methods.
For LLM latency, we compare the three repeats of each model, method, video and processing step, omitting an observation if its time is at least three times each of the other two original repeats. This rule is applied once to the original triplet, uniformly across methods.
For each input-length interval, we divide the sum of retained times by the corresponding total number of real original frames represented in those measurements.
For the separate aggregate Vision+LLM results, we retain actual KV lengths of at least 12K for LLaVA and 6K for Qwen, pooling measured time and real original-frame counts before taking their ratio.
Computation for padding is included, but duplicated padding frames are excluded from the denominator.
For the latency-growth summary, we fix the videos that reach 30K actual KV entries for LLaVA and 15K for Qwen under all compared methods: 19 OVO-Bench and 10 StreamingBench videos.
For this fixed set of videos, we compute $\mathrm{Lat}_{\mathrm{early}}$ and $\mathrm{Lat}_{\mathrm{late}}$ by pooling retained LLM times and represented real original-frame counts over 0--16K and 24K--end for LLaVA, and 0--8K and 12K--end for Qwen, respectively.
We report the observed increase as $100(\mathrm{Lat}_{\mathrm{late}}/\mathrm{Lat}_{\mathrm{early}}-1)$.
This summary uses the fixed set of videos, while the interval columns use all eligible videos; it is not computed from the displayed rounded interval means.
Refresh rate is the fraction of refresh decisions, excluding initial bank construction, that result in an actual bank refresh.

\paragraph{Implementation and measurement environment.}
Main accuracy and latency measurements use an NVIDIA H100 80GB with BF16.
Latency includes candidate management, bank refreshes, ReKV's internal KV transfers, and waiting for operation completion.
Video decoding, preprocessing, input transfer, initial text-prefix processing, question processing, and answer generation are excluded from the measured scope.
ReKV's question-time retrieval is also outside this scope.
Component profiling is conducted separately from ordinary latency measurement.

For Qwen, {\name} at $\tau=0.5$ was measured in a follow-up session using the frozen inputs and implementation; its {\Dense}, {\ExactTopK}, and ReKV references come from earlier measurements. Session differences remain a limitation.
Validated replacement measurements are used for cases flagged during timing review.

\begin{table*}[t]
\centering
\begingroup
\resulttablefont
\setlength{\tabcolsep}{2pt}
\renewcommand{\arraystretch}{1.04}
\begin{tabularx}{\textwidth}{@{}l*{5}{>{\raggedleft\arraybackslash}X}>{\raggedleft\arraybackslash}p{42pt}@{\hspace{6pt}}*{5}{>{\raggedleft\arraybackslash}X}>{\raggedleft\arraybackslash}p{42pt}@{}}
\toprule
 & \multicolumn{6}{c}{\textbf{LLaVA-OV-0.5B}} & \multicolumn{6}{c}{\textbf{Qwen2.5-VL-3B}} \\
\cmidrule(lr){2-7}\cmidrule(lr){8-13}
\textbf{Method} & \mbox{0--8K} & \mbox{8--16K} & \mbox{16--24K} & \mbox{24--30K} & \mbox{30K--end} & \shortstack{Growth\\(\%) $\downarrow$} & \mbox{0--4K} & \mbox{4--8K} & \mbox{8--12K} & \mbox{12--15K} & \mbox{15K--end} & \shortstack{Growth\\(\%) $\downarrow$} \\
\midrule
\multicolumn{13}{@{}l@{}}{\textbf{OVO-Bench}} \\
\textbf{{\Dense}} & 49.68 & 55.63 & 68.43 & 82.60 & 97.05 & +63.2 & 36.37 & 36.13 & 43.32 & 52.92 & 61.75 & +48.7 \\
\textbf{{\ExactTopK}} & 16.34 & 21.42 & 29.95 & 36.35 & 40.93 & +99.1 & 11.83 & 11.80 & 13.73 & 16.24 & 18.69 & +38.9 \\
\textbf{ReKV} & 25.76 & 28.94 & 28.99 & 29.02 & 29.00 & +5.3 & 26.23 & 25.83 & 26.21 & 26.55 & 62.69 & +27.2 \\
\rowcolor{paperTableHighlight}
\textbf{{\name}} & 18.79 & 18.56 & 18.97 & 19.50 & 19.80 & +4.7 & 15.89 & 15.88 & 16.22 & 16.61 & 17.49 & +3.7 \\
\midrule
\multicolumn{13}{@{}l@{}}{\textbf{StreamingBench}} \\
\textbf{{\Dense}} & 50.55 & 55.86 & 69.10 & 83.35 & 99.04 & +62.9 & 35.91 & 35.79 & 43.49 & 53.70 & 65.51 & +53.1 \\
\textbf{{\ExactTopK}} & 17.05 & 21.87 & 30.74 & 36.74 & 40.47 & +96.0 & 11.67 & 11.69 & 13.77 & 16.46 & 19.80 & +43.2 \\
\textbf{ReKV} & 26.46 & 29.19 & 29.19 & 29.18 & 29.00 & +3.5 & 25.83 & 25.46 & 25.85 & 26.52 & 66.34 & +28.8 \\
\rowcolor{paperTableHighlight}
\textbf{{\name}} & 19.26 & 18.79 & 19.21 & 19.61 & 19.57 & +2.7 & 15.95 & 15.99 & 16.33 & 16.94 & 18.55 & +4.9 \\
\bottomrule
\end{tabularx}
\endgroup

\caption{\textbf{LLM latency} (ms per original frame; lower is better). Columns show each model's actual KV length (K = 1,000 entries). Growth compares early and late frame-weighted means on a fixed set of videos: 0--16K versus 24K--end for LLaVA, and 0--8K versus 12K--end for Qwen. Aggregation is defined in Sec.~\ref{sec:setup}.}
\label{tab:efficiency}
\end{table*}

\section{Results and Discussion}
\label{sec:results}

We first compare latency and answer accuracy under the same main configurations, then examine the refresh policy and its costs.
The main configuration uses $\tau=0.5$ for Qwen and $\tau=0$ for LLaVA. The threshold and refresh-policy ablations are separate experiments.

\subsection{Latency Analysis}
\label{sec:main}
\label{sec:efficiency}

\label{sec:input_length}
We compare LLM-prefill latency per original frame across models and datasets.
{\name} achieves lower mean latency than ReKV in all measured length bins for both models and benchmarks, as reported in \cref{tab:efficiency}.

A key characteristic of {\name} is the small increase in per-frame latency as the KV history grows.
For a fixed set of videos, we compare frame-weighted mean latencies over early and late actual KV intervals: 0--16K versus 24K--end for LLaVA, and 0--8K versus 12K--end for Qwen.
{\name}'s latency increases by 2.7--4.7\% for LLaVA and 3.7--4.9\% for Qwen.
In the same comparison, {\ExactTopK} increases by 96.0--99.1\% and 38.9--43.2\%, respectively.
These results indicate that {\name}'s per-frame latency is less sensitive to history length within the evaluated video population and length range.

The length ranges in which {\name} achieves lower absolute latency than {\ExactTopK} vary by model.
For LLaVA, {\name} is faster in every measured bin from 8K onward, while for Qwen its advantage appears in the longest, 15K--end actual KV interval.
These results highlight the importance of considering both short-input latency and latency growth as frames accumulate in streaming prefill that retains the complete history.

Including vision encoding over actual KV lengths of at least 12K for LLaVA and 6K for Qwen, the relative performance remains model-dependent.
On LLaVA, {\name} takes 25.42 and 25.77\,ms per original frame on OVO and StreamingBench, respectively, compared with 35.66/36.35\,ms for {\ExactTopK} and 36.85/37.13\,ms for ReKV.
These correspond to latency reductions of 28.7\%/29.1\% relative to {\ExactTopK} and 31.0\%/30.6\% relative to ReKV, in OVO/StreamingBench order.
On Qwen, {\name} takes 41.69/41.95\,ms, compared with 39.25/39.97\,ms for {\ExactTopK} and 54.18/53.79\,ms for ReKV.
Its Vision+LLM latency is therefore 23.0\%/22.0\% lower than ReKV and 6.2\%/5.0\% higher than {\ExactTopK}.

\subsection{Accuracy Analysis}
\label{sec:accuracy}
We compare {\name}'s accuracy with the baselines.
We evaluate task-level accuracy and question-weighted overall accuracy using all 182 OVO and 87 StreamingBench questions; the results are summarized in \cref{tab:accuracy}.

\begin{table*}[t]
\centering
\begingroup
\resulttablefont
\setlength{\tabcolsep}{3pt}
\renewcommand{\arraystretch}{0.92}
\setlength{\aboverulesep}{1pt}
\setlength{\belowrulesep}{1.5pt}
\begin{tabularx}{\textwidth}{@{}ll*{10}{>{\raggedleft\arraybackslash}X}@{}}
\toprule
\multicolumn{12}{@{}l}{\textbf{OVO-Bench}} \\
\midrule
\textbf{Model} & \textbf{Method} & \textbf{OCR} & \textbf{ACR} & \textbf{ATR} & \textbf{STU} & \textbf{FPD} & \textbf{OJR} & \textbf{EPM} & \textbf{ASI} & \textbf{HLD} & \textbf{Overall} \\
\midrule
\multirow{4}{*}{\textbf{LLaVA-OV 0.5B}} & {\Dense} & 60.00 & 61.90 & 46.15 & 33.33 & 57.14 & 32.14 & 46.15 & 27.27 & 100.00 & 45.05 \\
 & {\ExactTopK} & 60.00 & 61.90 & 38.46 & 33.33 & 61.90 & 32.14 & 46.15 & 22.73 & 100.00 & 44.51 \\
 & ReKV & 50.00 & 57.14 & 30.77 & 30.30 & 61.90 & 35.71 & 53.85 & 22.73 & 100.00 & 42.31 \\
 & \cellcolor{paperTableHighlight}{\name} & \cellcolor{paperTableHighlight}63.33 & \cellcolor{paperTableHighlight}61.90 & \cellcolor{paperTableHighlight}46.15 & \cellcolor{paperTableHighlight}39.39 & \cellcolor{paperTableHighlight}61.90 & \cellcolor{paperTableHighlight}35.71 & \cellcolor{paperTableHighlight}61.54 & \cellcolor{paperTableHighlight}27.27 & \cellcolor{paperTableHighlight}100.00 & \cellcolor{paperTableHighlight}48.90 \\
\midrule
\multirow{4}{*}{\textbf{Qwen2.5-VL 3B}} & {\Dense} & 83.33 & 66.67 & 92.31 & 57.58 & 66.67 & 71.43 & 69.23 & 68.18 & 0.00 & 70.33 \\
 & {\ExactTopK} & 83.33 & 66.67 & 84.62 & 54.55 & 61.90 & 71.43 & 61.54 & 68.18 & 0.00 & 68.13 \\
 & ReKV & 83.33 & 66.67 & 92.31 & 57.58 & 66.67 & 71.43 & 69.23 & 68.18 & 0.00 & 70.33 \\
 & \cellcolor{paperTableHighlight}{\name} & \cellcolor{paperTableHighlight}83.33 & \cellcolor{paperTableHighlight}57.14 & \cellcolor{paperTableHighlight}84.62 & \cellcolor{paperTableHighlight}57.58 & \cellcolor{paperTableHighlight}66.67 & \cellcolor{paperTableHighlight}67.86 & \cellcolor{paperTableHighlight}61.54 & \cellcolor{paperTableHighlight}77.27 & \cellcolor{paperTableHighlight}0.00 & \cellcolor{paperTableHighlight}68.68 \\
\bottomrule
\end{tabularx}
\par\vspace{2pt}
\begin{tabularx}{\textwidth}{@{}ll*{11}{>{\raggedleft\arraybackslash}X}@{}}
\toprule
\multicolumn{13}{@{}l}{\textbf{StreamingBench}} \\
\midrule
\textbf{Model} & \textbf{Method} & \textbf{CS} & \textbf{OR} & \textbf{AR} & \textbf{PR} & \textbf{ActR} & \textbf{SU} & \textbf{EU} & \textbf{CT} & \textbf{TR} & \textbf{CR} & \textbf{Overall} \\
\midrule
\multirow{4}{*}{\textbf{LLaVA-OV 0.5B}} & {\Dense} & 37.50 & 80.00 & 75.00 & 100.00 & 77.78 & 25.00 & 66.67 & 31.25 & 60.00 & 71.43 & 62.07 \\
 & {\ExactTopK} & 37.50 & 80.00 & 68.75 & 100.00 & 66.67 & 25.00 & 66.67 & 25.00 & 60.00 & 42.86 & 56.32 \\
 & ReKV & 37.50 & 86.67 & 62.50 & 50.00 & 77.78 & 25.00 & 66.67 & 25.00 & 100.00 & 57.14 & 58.62 \\
 & \cellcolor{paperTableHighlight}{\name} & \cellcolor{paperTableHighlight}37.50 & \cellcolor{paperTableHighlight}73.33 & \cellcolor{paperTableHighlight}75.00 & \cellcolor{paperTableHighlight}75.00 & \cellcolor{paperTableHighlight}88.89 & \cellcolor{paperTableHighlight}25.00 & \cellcolor{paperTableHighlight}100.00 & \cellcolor{paperTableHighlight}43.75 & \cellcolor{paperTableHighlight}40.00 & \cellcolor{paperTableHighlight}57.14 & \cellcolor{paperTableHighlight}62.07 \\
\midrule
\multirow{4}{*}{\textbf{Qwen2.5-VL 3B}} & {\Dense} & 62.50 & 73.33 & 100.00 & 75.00 & 77.78 & 50.00 & 100.00 & 68.75 & 80.00 & 85.71 & 78.16 \\
 & {\ExactTopK} & 50.00 & 73.33 & 93.75 & 75.00 & 66.67 & 100.00 & 100.00 & 62.50 & 80.00 & 85.71 & 75.86 \\
 & ReKV & 62.50 & 73.33 & 100.00 & 75.00 & 77.78 & 75.00 & 100.00 & 68.75 & 80.00 & 85.71 & 79.31 \\
 & \cellcolor{paperTableHighlight}{\name} & \cellcolor{paperTableHighlight}50.00 & \cellcolor{paperTableHighlight}73.33 & \cellcolor{paperTableHighlight}93.75 & \cellcolor{paperTableHighlight}75.00 & \cellcolor{paperTableHighlight}66.67 & \cellcolor{paperTableHighlight}100.00 & \cellcolor{paperTableHighlight}100.00 & \cellcolor{paperTableHighlight}56.25 & \cellcolor{paperTableHighlight}80.00 & \cellcolor{paperTableHighlight}85.71 & \cellcolor{paperTableHighlight}74.71 \\
\bottomrule
\end{tabularx}
\endgroup

\caption{\textbf{Task-level multiple-choice accuracy} (\%; higher is better). Overall is question-weighted. OVO-Bench: OCR (optical character recognition), ACR (action recognition), ATR (attribute recognition), STU (spatial understanding), FPD (future prediction), OJR (object recognition), EPM (episodic memory), ASI (action sequence identification), and HLD (hallucination detection; one question). StreamingBench: CS (clips summarization), OR (object recognition), AR (attribute recognition), PR (prospective reasoning), ActR (action recognition), SU (spatial understanding), EU (event understanding), CT (counting), TR (text-rich understanding), and CR (causal reasoning).}
\label{tab:accuracy}
\label{tab:accuracy_by_task}
\end{table*}

On LLaVA, {\name} obtains 48.90\% on OVO and 62.07\% on StreamingBench, compared with 44.51\% and 56.32\% for {\ExactTopK}.
Its OVO accuracy is higher in six of the nine represented tasks and equal in the other three.
On StreamingBench, gains occur in five tasks, including the object and event counting task (CT; 43.75\% versus 25.00\%) and action recognition (88.89\% versus 66.67\%), with ties in two tasks and declines in three.
On Qwen, {\name} obtains 68.68\% on OVO and 74.71\% on StreamingBench, compared with 68.13\% and 75.86\% for {\ExactTopK}.
The Qwen StreamingBench task scores match {\ExactTopK} in nine of ten tasks; one fewer correct CT answer (56.25\% versus 62.50\%) accounts for the 1.15-percentage-point overall gap.
The separate threshold experiment in \cref{sec:ablation} examines refresh-policy sensitivity on questions that include evidence accumulated across the video.
The main task breakdown and this separate ablation use their respective recorded predictions; the ablation does not replace the main results.

\subsection{Ablation Study and Discussion}
\label{sec:ablation}
\label{sec:analysis}

\begin{figure}[!htb]
\centering
\includegraphics{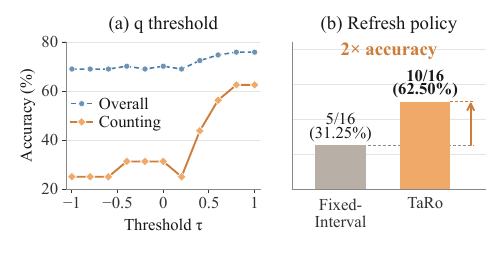}
\caption{Qwen refresh-policy ablation on StreamingBench. (a) Accuracy versus threshold. (b) TaRo versus Fixed-Interval on CT at similar observed refresh rates; Fixed-Interval rebuilds banks every 14 original frames (7 processing steps).}
\label{fig:qwen-refresh-comparison}
\end{figure}

\paragraph{Q threshold.}
The refresh rule is $\cos(q,q^{\mathrm{ref}})<\tau$ or the age bound: lowering $\tau$ makes similarity-triggered refresh less likely.
We vary $\tau$ for Qwen2.5-VL-3B on our StreamingBench evaluation subset.
Our CT subset contains 16 questions from 15 videos, primarily asking for cumulative event counts, alongside object-counting questions.
Many questions require evidence accumulated across earlier frames, making this subset useful for examining the effect of bank refresh.
Increasing $\tau$ from 0 to 0.5 doubles CT accuracy from 31.25\% to 62.50\%, as shown in Fig.~\ref{fig:qwen-refresh-comparison}(a).
We examine the history-dependent cases in the comparison with Fixed-Interval Refresh below.

\paragraph{Adaptive versus Fixed-Interval Refresh.}
We compare query-adaptive {\name} at $\tau=0.5$ with Fixed-Interval Refresh on the same 16-question CT subset.
Fixed-Interval Refresh rebuilds candidate banks every 14 original frames (7 processing steps), while retaining the same candidate selection and attention procedure as {\name}.
Comparing policies at similar refresh rates helps control the frequency of full-history searches, allowing us to assess the benefit of query-guided refresh timing.

{\name} and Fixed-Interval Refresh have observed refresh rates of 14.58\% and 13.83\%, respectively.
At these similar refresh frequencies, {\name} achieves $2\times$ the CT accuracy of Fixed-Interval Refresh: 62.50\% versus 31.25\%, as shown in Fig.~\ref{fig:qwen-refresh-comparison}(b).
These results highlight the value of query-guided refresh for questions that rely on visual evidence accumulated across earlier frames.

\paragraph{Limitations.}
{\name} retains the full KV history and therefore does not reduce KV-cache memory growth, while candidate banks and reference queries require additional storage.
Our evaluation focuses on video prefill and does not measure end-to-end response latency including question processing and answer generation.
The experiments cover actual KV lengths up to approximately 32K for LLaVA and 16K for Qwen, and performance at longer contexts requires further evaluation.

\section{Conclusion}
\label{sec:conclusion}

We presented {\name}, which reduces repeated full-history search in streaming video prefill through position-wise KV candidate reuse and refresh guided by query changes and bank age.
Across two models and two benchmarks, {\name} achieves $1.37$--$3.58\times$ LLM-prefill speedups over ReKV.
On a fixed set of videos, its latency grows by less than 5\% between early and late history intervals, compared with 39--99\% for {\ExactTopK}.
These results highlight the value of separating candidate discovery from per-frame selection, allowing attention to adapt to the current query while limiting latency growth as the KV history accumulates.

\section*{LLM Usage Statement}
\label{sec:llm_usage}

We used OpenAI's ChatGPT and Codex to assist with code development, literature review, and manuscript preparation, including visual materials. The authors take full responsibility for the scientific content and final manuscript.

{
    \small
    \bibliographystyle{ieeenat_fullname}
    \bibliography{main}
}
\clearpage
\EnableSupplementaryFloats
\ifdefined\XeTeXversion\fontencoding{T1}\selectfont\fi
\maketitlesupplementary
\appendix
\providecommand{\mainsecref}[1]{\ref{#1}}

\begin{figure*}[!b]
  \centering
  \includegraphics[width=0.96\textwidth]{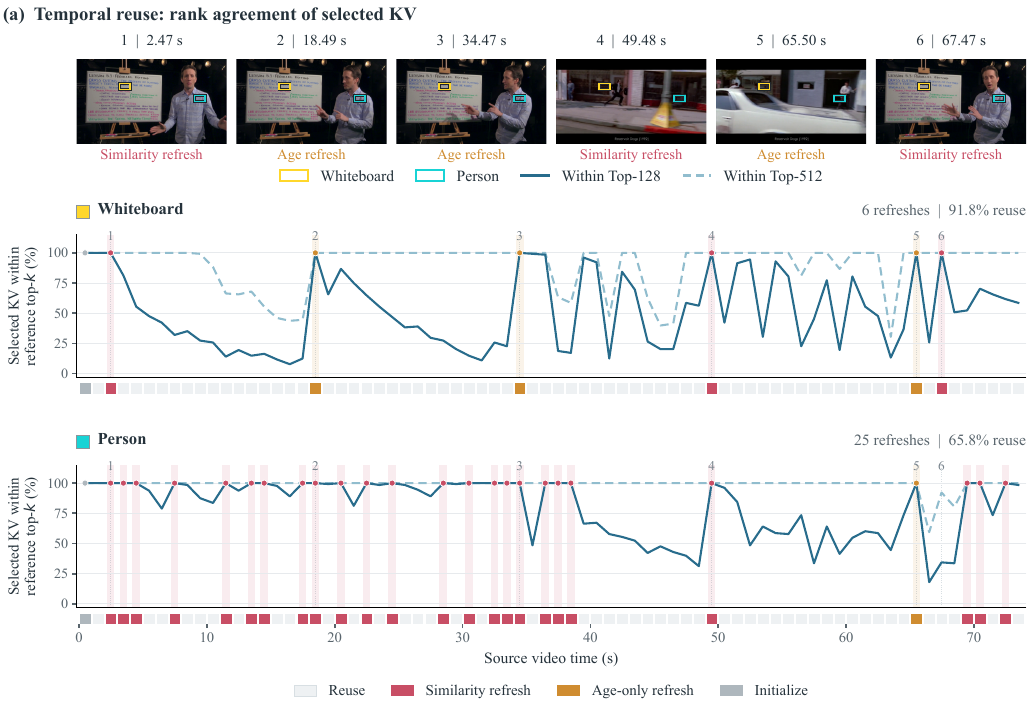}
  \caption{\textbf{Candidate reuse in time and space.}
  \textbf{(a)} Selected KV within exact full-causal Top-128 (solid) and Top-512 (dashed).
  Colored strips identify the executed refresh decisions.}
  \label{fig:supp_candidate_reuse}
\end{figure*}
\section{Temporal Candidate Reuse}

\noindent
{\name} decides whether to refresh candidate banks at each spatial position using
query similarity and bank age (main paper, Sec.~\mainsecref{sec:refresh}). Figure~\ref{fig:supp_candidate_reuse}(a) compares two
fixed positions in a StreamingBench video (\texttt{sb\_93\_1}): the whiteboard and
the presenter. We use Qwen2.5-VL-3B, layer 19, head 1, $k=128$, $M=512$,
$\tau=0.5$, and a 16-step age cap.

\smallskip
\noindent\textbf{Static and dynamic regions.}
After the first similarity refresh at 2.47\,s, the whiteboard position remains
stable during the lecture: its only refreshes before the scene change are
age-only events at 18.49 and 34.47\,s. The presenter position instead refreshes
frequently as the person moves. Both positions trigger similarity refreshes
when the inserted film begins at 49.48\,s.
\newpage
\noindent During this inserted segment, both
positions then reuse their banks until the age-only refresh at 65.50\,s;
the whiteboard position refreshes by similarity again when the lecture returns
at 67.47\,s. The boxes mark fixed image coordinates, so their content changes
across cuts; the presenter location also becomes more stable in the inserted scene.

\smallskip
\noindent\textbf{Why include Top-512?}
Top-128 membership can change even when selected KV remain highly ranked.
The broader Top-512 reference shows this rank proximity using the same current
query and full causal cache. Both curves divide by the actual number of selected
KV; the attention budget remains 128. The dashed curve is a rank diagnostic,
not an accuracy or error bound.

\clearpage

\begin{figure*}[!b]
  \ContinuedFloat
  \centering
  \includegraphics[width=\textwidth]{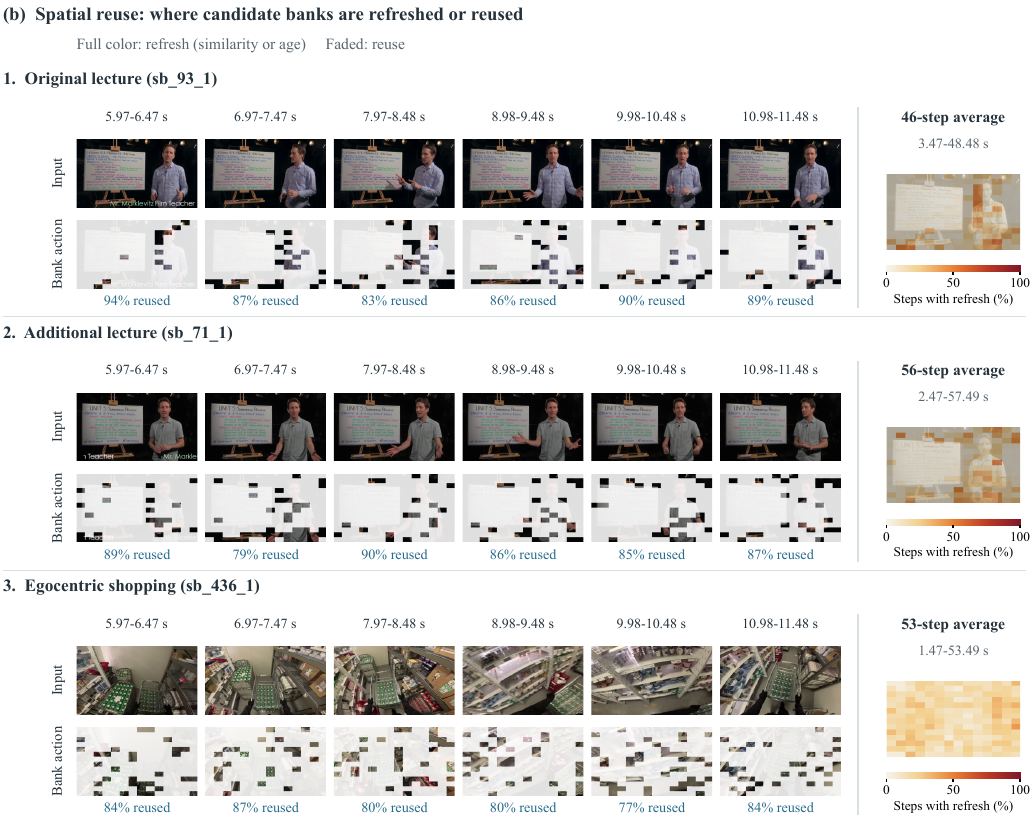}
  \caption{\textbf{Candidate reuse in time and space (continued).}
  \textbf{(b)} Six consecutive processing steps per video. Full color: refresh
  (similarity or age); faded: reuse. Average maps use all positions
  and steps in the stated interval, on the same 0--100\% scale.}
  \label{fig:supp_spatial_reuse}
  \vspace{6\baselineskip}
\end{figure*}
\section{Spatial Candidate Reuse}

\noindent
Figure~\ref{fig:supp_spatial_reuse}(b) compares two lecture videos and an egocentric
shopping video using the refresh policy in Sec.~\mainsecref{sec:refresh} of the main
paper, at layer 19 and GQA group 1.

The full-interval average maps show a clear spatial contrast. Refreshes are
concentrated around the moving presenters in the first two examples, but spread
across the image in the third.
\newpage
\noindent The static whiteboard regions are largely reused,
whereas camera motion in the shopping video changes the content at many image
positions. Even in this moving-camera example, 81.7\% of banks are reused per
step on average over the 53 noninitial steps.

\clearpage
\section{Algorithm and Reproducibility Details}

\noindent
Algorithm~\ref{alg:supp_taro} summarizes the bank construction, reuse, and refresh
in Secs.~\mainsecref{sec:bank}--\mainsecref{sec:refresh} of the main paper for one
query-head group $\mathcal G$ at a fixed layer and spatial position $p$. The step index $t$ counts one processed frame for
LLaVA and a frame pair for Qwen. Given $P$ current visual tokens and total cache
length $T_t$, let $\mathcal A=\{0,\ldots,T_t-P+p\}$ be the full causal prefix
and $\mathcal F=\{T_t-P,\ldots,T_t-P+p\}$ its current-step entries.
$\Top{b}$ returns the indices of the $\min(b,n)$ highest among $n$ valid scores.

\begin{algorithm}[H]
\caption{{\name} attention for one group $\mathcal G$ at $(l,p)$.}
\label{alg:supp_taro}
\small
\begin{algorithmic}[1]
\Require current step $t$; post-RoPE queries $q_h$, $h\in\mathcal G$; shared KV cache; causal sets $\mathcal A,\mathcal F$
\Require budgets $1\le k\le M$; threshold $\tau$; age bound $N_{\max}$ in processing steps
\Require saved per-head banks $\mathcal B_h$, reference queries $q_h^{\rm ref}$, and shared refresh step $t_r$, if initialized
\Ensure outputs $o_h$ and updated bank state
\State $r\gets\text{banks are uninitialized}$
\If{not $r$}
  \State $r\gets[\min_{h\in\mathcal G}\cos(q_h,q_h^{\rm ref})<\tau]$
  \State $r\gets r\ \lor\ [t-t_r\ge N_{\max}]$
\EndIf
\For{$h\in\mathcal G$}
  \State $\mathcal C_h\gets\mathcal A$ \textbf{if} $r$ \textbf{else} $\mathcal B_h\cup\mathcal F$
  \State $s_{h,j}\gets q_h^{\top}k_j/\sqrt d$, for $j\in\mathcal C_h$
  \State $\mathcal S_h\gets\Top{k}\{s_{h,j}:j\in\mathcal C_h\}$
  \State $o_h\gets\sum_{j\in\mathcal S_h}\softmax(s_{h,\mathcal S_h})_j v_j$
  \If{$r$}
    \State $\mathcal B_h\gets\Top{M}\{s_{h,j}:j\in\mathcal C_h\}$
    \State $q_h^{\rm ref}\gets q_h$
  \EndIf
\EndFor
\If{$r$}
  \State $t_r\gets t$
\EndIf
\State \Return $\{o_h\}_{h\in\mathcal G}$ and saved bank state
\end{algorithmic}
\end{algorithm}

\newpage
\noindent\textbf{Processing and refresh units.}
Following Sec.~\mainsecref{sec:setup} of the main paper, both models use 2 FPS,
BF16, batch size 1, $k=128$, and $M=512$.
The same 32-original-frame age cap corresponds to 16 seconds at the target sampling rate.
GQA heads share the refresh decision and age, but keep separate banks and reference queries.
Only video-prefill attention uses {\name}; question and answer tokens use {\Dense}.

\medskip
{\centering
\captionsetup{hypcap=false,skip=6pt}
\captionof{table}{\textbf{Model-specific processing and refresh settings.}}
\label{tab:supp_processing}
\renewcommand{\arraystretch}{1.05}
\small
\begin{tabularx}{\columnwidth}{@{}Xcc@{}}
\toprule
\textbf{Setting} & \shortstack{\textbf{LLaVA-OV}\\\textbf{0.5B}} & \shortstack{\textbf{Qwen2.5-VL}\\\textbf{3B}} \\
\midrule
Original frames per processing step & 1 & 2 \\
Visual tokens $P$ per processing step & 196 & 196 \\
Query heads / KV heads & 14 / 2 & 16 / 2 \\
Head dimension $d$ & 64 & 128 \\
Main similarity threshold $\tau$ & 0 & 0.5 \\
Maximum age $N_{\max}$ (processing steps) & 32 & 16 \\
Query chunk size & 16 & 16 \\
\bottomrule
\end{tabularx}\par}

\end{document}